\documentclass[letterpaper]{article} 
\usepackage[draft]{aaai2026}  
\usepackage{times}  
\usepackage{helvet}  
\usepackage{courier}  
\usepackage[hyphens]{url}  
\usepackage{graphicx} 
\usepackage{natbib}  
\usepackage{caption} 
\usepackage{algorithm}
\usepackage{algorithmic}
\usepackage{xspace}
\usepackage{amsmath}
\usepackage{amssymb}
\usepackage[capitalize,noabbrev]{cleveref}
\usepackage{adjustbox}
\usepackage{booktabs}       
\usepackage{multirow}
\usepackage{bm}
\usepackage{pifont}
\usepackage[switch]{lineno}

\def\onedot{.\xspace}
\def\eg{\emph{e.g}\onedot} 
\def\ie{\emph{i.e}\onedot}

\def\cmark{\ding{51}}
\def\xmark{\ding{55}}

\def\Ours{Knowledge Preservation and Unification\xspace}
\def\ours{KPU\xspace}

\newif\ifshowcmts
\showcmtsfalse

\usepackage{newfloat}
\usepackage{listings}
\DeclareCaptionStyle{ruled}{labelfont=normalfont,labelsep=colon,strut=off} 
\floatstyle{ruled}
\newfloat{listing}{tb}{lst}{}
\floatname{listing}{Listing}
\title{Seeing Further on the Shoulders of Giants: \\ Knowledge Inheritance for Vision Foundation Models}
\author{
    Jiabo Huang, Chen Chen, Lingjuan Lyu
}
\affiliations{
    Sony AI \\
}

\begin{document}

\maketitle

\begin{abstract}
Vision foundation models (VFMs) are predominantly developed using data-centric methods. 
These methods require training on vast amounts of data usually with high-quality labels, which poses a bottleneck for most institutions that lack both large-scale data and high-end GPUs. 
On the other hand, many open-source vision models have been pretrained on domain-specific data, 
enabling them to distill and represent core knowledge in a form that is transferable across diverse applications. 
Even though these
models are highly valuable assets, 
they remain largely under-explored in empowering the development of a general-purpose VFM.
In this paper, 
we present a new model-driven approach for training VFMs through joint knowledge transfer and preservation. 
Our method 
unifies multiple pre-trained teacher models in a shared latent space 
to mitigate the ``imbalanced transfer'' issue 
caused by their distributional gaps.
Besides,
we introduce a knowledge preservation strategy
to take a general-purpose teacher as 
a knowledge base
for integrating knowledge from the remaining purpose-specific teachers
using an adapter module.
By unifying and aggregating existing models,
we build a powerful VFM to inherit teachers' expertise
without needing to train on a large amount of labeled data.
Our model not only provides generalizable visual features,
but also inherently supports multiple downstream tasks.
Extensive experiments demonstrate that our VFM outperforms existing data-centric models across four fundamental vision tasks,
including image classification, object detection, semantic and instance segmentation.
\end{abstract}

\section{Introduction}

On the path toward achieving artificial general intelligence (AGI), 
the development of AI systems capable of handling various tasks simultaneously 
with a comprehensive understanding of data has become a pressing topic across multiple 
communities~\cite{brown2020gpt2,yuan2021florence,lu2022unified-io}. 
In the computer vision domain, there is a growing interest in training foundation models that can encode diverse levels of visual knowledge and seamlessly adapt to different downstream tasks with minimal effort~\cite{zhu2022uniperceiver,li2023uniperceiver2,xiao2023florence2}.

\begin{figure*}[t]
\centering
\includegraphics[width=\linewidth]{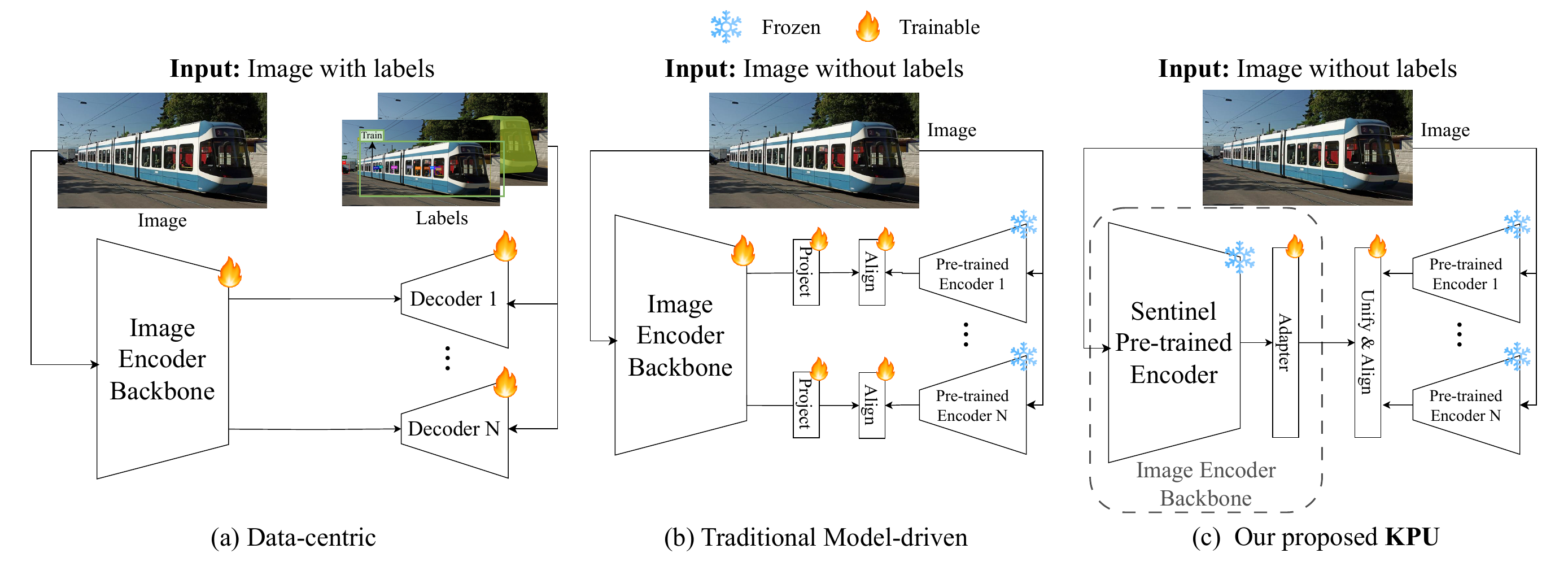}
\caption{
Training paradigms for Vision Foundation Models. 
(a) Most of the data-centric models, which can be used out of the box, 
are trained based on large-scale labeled data~\cite{lu2023unified-io2}.
(b) Existing model-driven approaches inherit pre-learned visual knowledge from diverse teachers
by feature alignment on their latent spaces~\cite{Ranzinger_2024_CVPR}.
(c) Our
proposed \Ours (\textbf{\ours})
method
combines knowledge preservation and unified transfer
to aggregate complementary insights from multiple pre-trained models
upon a general-purpose sentinel teacher.
}
\label{fig:teaser}
\end{figure*}

Vision Foundation Models (VFMs) can be developed using a variety of techniques, 
broadly categorized into data-centric and model-driven approaches based on their distinct sources of knowledge. 
Data-centric methods~\cite{oquab2023dinov2,he2022mae,mizrahi20244m,bachmann20244m} (\cref{fig:teaser}(a))
derive either visual features from large-scale unlabeled data~\cite{he2020momentum,chen2020simple} 
or task-aware predictions from accurate data-label correspondences~\cite{yuan2021florence,xiao2023florence2,lu2023unified-io2}.
While the former unsupervised representation learning methods
can barely support any downstream tasks without further fine-tuning with labels, 
the latter supervised approaches
face challenges in generalization 
due to the prohibitive cost of scaling up high-quality labels for all tasks of interest.
In contrast, 
model-driven methods~\cite{Ranzinger_2024_CVPR,wang2024sam,ypsilantis2024udon,yao2023moma,sariyildiz2024unic} 
(\cref{fig:teaser}(b)) 
represent a parallel paradigm for building VFMs by 
aggregating pre-learned knowledge from existing models 
rather than relying on labeled data. 
These approaches aim at training
a versatile student model by mimicking the behaviors of diverse teachers in their latent spaces. 
However,
the distinct training schemes adopted by teachers usually result in unique feature distributions
with significant variations in magnitudes~\cite{heinrich2025radiov2}.
In this case,
teachers are implicitly weighted
when training with commonly adopted feature matching losses
that are sensitive to variance scale~\cite{ranzinger2024phi},
\eg, Mean Squared Error (MSE) or Smooth-L1.
Such uncontrolled weighting is likely to result in
teacher dominance during the learning process.
This will 
intuitively limit the generalization ability of students
by being biased to dominated teachers
instead of exploring the synergies of all teachers~\cite{caruana1997multitask,zamir2018taskonomy},
\ie, imbalanced transfer.


In this work, 
we introduce a new model-driven method for training vision foundation models
by \Ours (\textbf{\ours}).
As illustrated in \cref{fig:teaser} (c),
\ours adopts the visual knowledge derived by multiple off-the-shelf models 
and explores their complementary insights in a common latent space. 
Different from existing model-driven methods that 
solely align the student with every teacher in independent latent spaces (\cref{fig:teaser} (b)),
we project all the teachers into a shared space for their alignment via the student.
Such a unification design helps with 
shrinking the gaps among teacher distributions
to mitigate the impacts of imbalanced transfer~\cite{sariyildiz2024unic,ranzinger2024phi,Ranzinger_2024_CVPR}.
In addition,
we introduce a knowledge preservation strategy
to retain the knowledge from a general-purpose teacher as the sentinel.
The frozen sentinel teacher aggregates diverse knowledge from the remaining purpose-specific teachers with an additional trainable adapter.
This strategy allows us to benefit from 
the remarkable generalization capability of general-purpose teachers
by explicitly reusing their well-trained weights.
Meanwhile,
we also acquire task-aware predictive abilities
from purpose-specific teachers to support multiple downstream tasks
without requiring labeled data for further adaptation.

In this paper, we make three contributions.
\textbf{(1)}
We propose a simple yet effective model-driven training method for VFMs
that combines knowledge preservation and unification.
Our approach facilitates effective knowledge transfer
by integrating diverse teacher knowledge into a general knowledge base
while mitigating imbalanced transfer by teacher unification.
\textbf{(2)}
We build a VFM using the proposed training method,
which provides expressive visual feature representations
to benefit various downstream tasks.
Besides, our model inherits zero-shot classification and object detection capabilities 
from its teachers, without relying on any task-specific labeled data.
\textbf{(3)}
We highlight the potential of model-driven VFMs 
by demonstrating superior performance over extensive data-centric models
across four fundamental vision tasks. 
For example,
with around $10\%$ of the training data used in DINOv2~\cite{oquab2023dinov2} and Florence-2~\cite{xiao2023florence2}, 
our model outperforms DINOv2 by $3.3\%$ on the downstream instance segmentation task
and Florence-2 by $41.5\%$ on the zero-shot object detection task
even without using any object bounding-box labels throughout our training process.

\section{Related Work}

\paragraph{Vision Foundation Models.} 
Vision Foundation Models (VFMs) aim to derive universal visual knowledge from large-scale data, 
enabling exceptional generalization across diverse downstream tasks~\cite{yuan2021florence}. 
Early VFMs focused on visual representation learning~\cite{oquab2023dinov2,he2022mae,chen2020simple,he2020momentum} by training general-purpose visual encoders that serve as the backbones for downstream task training. These methods are typically designed with minimal reliance on manual annotations.
However, since these backbone models are trained without incorporating task-aware information, 
they can only support limited downstream tasks unless further task-specific training is performed.
In contrast, inspired by the success of sequence-to-sequence modeling in natural language processing (NLP)~\cite{brown2020gpt2}, recent VFMs have shifted their focus from data generalization to task generalization~\cite{lu2022unified-io,lu2023unified-io2,li2023uniperceiver2,xiao2023florence2}. 
These approaches aim to re-formulate as many vision tasks as possible to share a common input-output structure, 
enabling simultaneous learning of all selected vision tasks using a unified sequence modeling architecture. 
While such purpose-specific VFMs can be directly 
used for many downstream tasks encountered during training or similar ones,
their reliance on the quantity and quality of task-specific labels
hampers their generalization ability to unseen data.

\paragraph{Knowledge Transfer.}
Knowledge Transfer is a widely studied technique that learns from the abstract knowledge already derived and encoded in off-the-shelf models~\cite{gou2021knowledge,Hinton2015DistillingTK}. 
Early methods in knowledge transfer focused on distilling knowledge from a large teacher model to a smaller student model by 
supervising student’s training based on teacher’s predictions in the target label space~\cite{Hinton2015DistillingTK,heo2019comprehensive,huang2017like,wei2022contrastive,zagoruyko2016paying}. 
%
Recently, knowledge transfer has been effectively applied to train VFMs, 
where the student model is designed to combine the advantages of multiple pre-trained teacher models~\cite{Ranzinger_2024_CVPR,wang2024sam,sariyildiz2024unic},
which can be VFMs themselves. 
Due to the heterogeneous nature of the label spaces associated with the different tasks addressed by different teacher models, 
these methods typically transfer knowledge through feature instead of prediction matching~\cite{Ranzinger_2024_CVPR,sariyildiz2024unic,heinrich2025radiov2} 
for simultaneous teacher-student alignment across distinct spaces.
However, given the differences in data domains and training objectives of teacher models, there is no guarantee that the 
training supervisions derived from aligning with different teachers are compatible~\cite{yu2020gradient}. 
To avoid imbalanced transfer and to better exploit the potential synergies among teachers,
several existing model-driven approaches apply
handcrafted distribution transforms or weighting~\cite{ranzinger2024phi,heinrich2025radiov2,sariyildiz2024unic}
for balancing the energy spent on learning from each teacher.
Instead of direct manipulation of teacher distributions,
our knowledge unification shrinks their gaps by aligning them
via the same student in a shared space.
Our experiments demonstrate
the effectiveness of such a learnable solution
regardless of its simplicity.

\section{\Ours
}
\begin{figure*}[t]
\centering
\includegraphics[width=\linewidth]{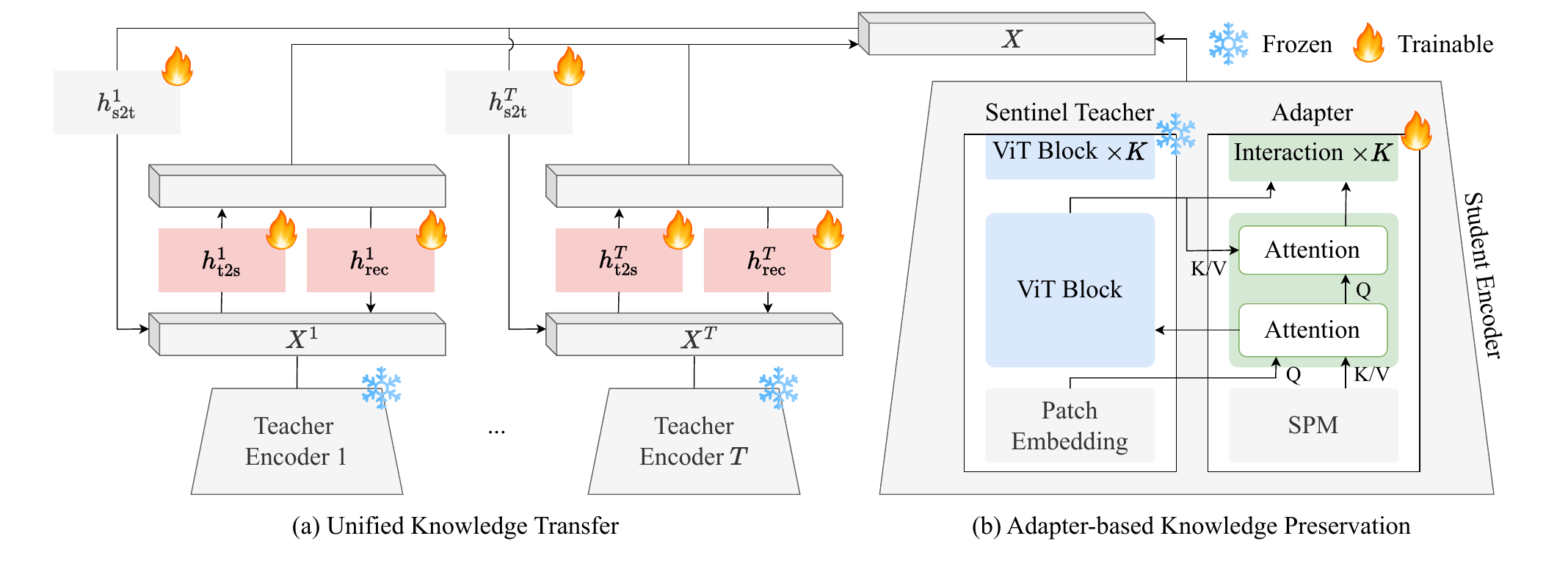}
\caption{
Overview of the proposed \ours method. 
(a) \textbf{Unified Knowledge Transfer}.
Knowledge from multiple pre-trained models is aggregated and transferred 
to the student model via feature alignment, 
performed not only in individual teachers’ latent spaces but also 
in a unified common latent space. 
(b) \textbf{Adapter-based Knowledge Preservation}.
We explicitly reuse and retain the pre-trained weights of a sentinel teacher 
while incorporating additional knowledge from other teachers through an additional trainable adapter~\cite{chen2022vision}.
}
\label{fig:overview}
\end{figure*}
Suppose we have $T$ well-trained teacher models $\{g_{\theta^t}\}_{t=1}^T$,
our knowledge inheritance objective 
is to aggregate their knowledge represented in their features of 
$N$ unlabeled images $\mathcal{I} = \{\bm{I}_1, \bm{I}_2, ..., \bm{I}_N\}$.
To this end,
we develop a student model $f_\theta$ parameterized by $\theta$,
which takes an image as input and
learns rich visual feature representations
$X=\{\bm{x},\bm{V}\}$, 
where $f_\theta: \bm{I} \rightarrow X$ and 
$\bm{x} \in \mathbb{R}^D$ is the $D$-dimensional image feature while 
$\bm{V} \in \mathbb{R}^{H\times W\times D}$ is the feature map of size $H\times W$.
To achieve comprehensive visual features that support diverse downstream tasks requiring various levels of visual understanding, 
we integrate teacher models trained with distinct focuses. 
These include CLIP~\cite{radford2021learning}, which provides global image matching supervision to language;
GroundingDINO~\cite{liu2023grounding}, a region-level object detection expert; 
and DINOv2~\cite{oquab2023dinov2}, deriving general visual knowledge purely from data
without being biased to any types of labels.
The diverse viewpoints offered by these teacher models result in feature representations that 
collectively serve as a more robust and comprehensive description of the imagery data.
Such diversity is essential to the development of a unified and strong student model. 
However, achieving this integration 
is challenging due to the potentially incompatible visual knowledge encoded in the latent spaces, 
as each teacher model is independently trained to satisfy distinct objective functions
on diverse data domains.

In this paper,
we propose \Ours (\ours)
for constructing vision foundation models by aggregating and inheriting knowledge from multiple heterogeneous teachers. 
An overview of \ours is depicted in \cref{fig:overview}.
\ours projects all teachers’ latent embeddings into a shared common space, 
enabling the exploration of their synergies 
by avoiding dominance caused by vastly different distributions.
Furthermore,
we introduce a knowledge preservation strategy
for effective knowledge aggregation
into a general-purpose sentinel teacher
by an additional adapter module.
The pre-learned weights of the sentinel teacher
provide a strong starting point for teacher aggregation,
while the additional knowledge injected by the adapter 
enables our model to be applied to downstream tasks
without further adaptation using labeled data.


\subsection{Unified Knowledge Transfer
}
To inherit the knowledge offered by off-the-shelf teacher models, 
we train the student model to mimic the behaviors of the teachers through feature alignment between each student-teacher pair. 
Specifically, given the features $X^t=g_{\theta^t}(\bm{I})=\{\bm{x}^t, \bm{V}^t\}$ produced by the $t$-th teacher model for the image $\bm{I}$, 
the student’s feature map $\bm{V}$ is first resized via bilinear interpolation to match the spatial dimensions of the teacher’s feature map $\bm{V}^t$. 
It is then projected together with the image feature $\bm{x}$ using independent multilayer perceptrons (MLP),
each with a single hidden layer,
to align the student's feature dimension with that of the teacher. 
Notably, as features from different teachers vary in size and dimension, 
the MLP projection layers applied to the student model are customized for each teacher. 
Once the student’s features are adapted to match the spatial size and feature dimension of the teacher’s features, 
knowledge transfer is achieved by maximizing their cosine similarity while minimizing their smooth L1 distance:
\begin{equation}
\begin{aligned}
\mathcal{L}_\text{s2t}(\bm{I}) =& \frac{1}{T}\sum_{t=1}^T \mathcal{L}_\text{align}(h_\text{s2t}^t(X), X^t),\quad \text{where} \\
\mathcal{L}_\text{align}(X_1, X_2) =& \mathcal{L}_\text{align}(\{\bm{x}_1, \bm{V}_1\}, \{\bm{x}_2, \bm{V}_2\}) \\
=& \lambda_1\mathcal{L}_\text{cos}(\bm{x}_1, \bm{x}_2) + \\
& \lambda_2\mathcal{L}_\text{cos}(\bm{V}_1, \bm{V}_2) + \\ 
& \lambda_3\mathcal{L}_\text{smooth-l1}(\bm{V}_1, \bm{V}_2).
\end{aligned}
\label{eq:kt_s2t}
\end{equation}
In \cref{eq:kt_s2t}, the overall alignment loss is the average of matching the student with each teacher, 
and $h^t_{s2t}(\bm{x})$ stands for the spatial resize and feature projection operations for aligning with the $t$-th teacher.
The hyperparameters $\{\lambda_i\}_{i=1}^3$ are used for balancing direction and intensity matching
in feature alignment,
which are set to $\lambda_1=1$, $\lambda_2=0.9$ and $\lambda_3=0.1$ following~\citet{Ranzinger_2024_CVPR}.
The first term in $\mathcal{L}_\text{align}$ is deprecated when aligning with teachers that don't produce an image-level global feature $\bm{x}^t$ to be matched with $\bm{x}$ from the student.

\paragraph{Knowledge unification.}
Considering that the feature matching losses in \cref{eq:kt_s2t}
are sensitive to the magnitude of teacher features,
aligning the student with each teacher in its own latent space
is prone to implicit teacher weighting that 
further leads to a student biased to the dominated teachers.
For better exploration of teachers' collective knowledge,
we propose to conduct student-teacher alignment
in a unified latent space shared by all teachers.
The formulation of learning from teachers' unification is similar to \cref{eq:kt_s2t} but the resize and projection operations $h_\text{t2s}^t$
are applied to the features of teachers rather than the student:
\begin{equation}
\mathcal{L}_\text{t2s}(\bm{I}) = \frac{1}{T}\sum_{t=1}^T\mathcal{L}_\text{align}(X, h^t_\text{t2s}(X^t)).
\label{eq:kt_t2s}
\end{equation}
Since both the student features and the projected teacher features are learnable, 
simply aligning them without additional constraints on what to be learned 
may result in information loss and a suboptimal solution.
In this case,
we introduce a simple reconstruction objective that projects the teacher’s features from the common space back into their original latent spaces by $h_\text{rec}^t$:
\begin{equation}
\mathcal{L}_\text{rec}(\bm{I}) = \frac{1}{T}\sum_{t=1}^T\mathcal{L}_\text{align}(X^t, h^t_\text{rec}(h^t_\text{t2s}(X^t))).
\label{eq:kt_rec}
\end{equation}
With such a regularization,
we construct the common space for not only feature alignment but also retaining the rich visual knowledge from teachers.

\subsection{Adapter-based Knowledge Preservation
}
Since the goal of training the student model is to develop a comprehensive understanding of visual data, 
and such knowledge is intrinsically encoded/distributed in the weights of teachers, 
it is intuitive to explicitly reuse and aggregate the teachers' pre-learned weights
for knowledge aggregation.
However, the diverse teacher models we are dealing with
are potentially built with distinct architectures and label spaces.
It is non-trivial to fuse the weights of such heterogeneous teachers.
To address this challenge, 
we formulate a hybrid strategy that combines knowledge transfer and preservation. 
Specifically, we reuse the frozen weights of a sentinel teacher while incorporating the knowledge from the remaining teachers through an additional trainable adapter.
Given the success of ViT-Adapter~\cite{chen2022vision} on
injecting the missing multiscale modeling capability
into the transformer architecture
for better support of vision tasks,
we propose to adopt it as an instantiation of our knowledge preservation.
Our adapter here aims at effective injection of diverse knowledge
from purpose-specific teachers
as a complementary remedy to the sentinel teacher which provides a general knowledge base.

\paragraph{Knowledge Adapter.}
As illustrated in \cref{fig:overview},
the adapter module adopts a convolutional spatial prior module (SPM)
and $K$ interaction blocks, 
each comprising an injector and an extractor, 
to enable bidirectional interaction between the adapter's features and the ViT features via cross-attentions~\cite{vaswani2017attention}. 
The SPM produces feature maps at multiple resolutions, 
supplying the ViT~\cite{dosovitskiy2020image} with rich visual information to enhance its patch embeddings. 
In turn, the updated patch tokens from the ViT blocks are utilized to 
refine the multiscale features generated by the adapter. 
This design leverages the robust feature representations of the ViT while 
enabling the adapter to focus on learning residual enhancements, 
thereby improving overall feature quality.
We refer to ViT-Adapter~\cite{chen2022vision} 
for more details about the implementation of SPM, injector and extractor.

\subsection{Model Training and Inference}
\paragraph{Training.}
Given the loss functions $\mathcal{L}_\text{t2s}$ and $\mathcal{L}_\text{s2t}$ for unified knowledge transfer through feature alignment, 
along with the regularization term to prevent 
information loss in the common latent space shared by the student and all the teachers, 
the overall training objective of \ours is defined as:  
\begin{equation}
\mathcal{L}_\text{\ours} = \mathcal{L}_\text{t2s} + \mathcal{L}_\text{s2t} + \lambda \mathcal{L}_\text{rec}.
\label{eq:loss_all}
\end{equation}
The $\lambda$ in \cref{eq:loss_all} is for balancing the knowledge transfer and regularization losses,
which is set to $1$ in practice.
%
For knowledge preservation,
we take DINOv2~\cite{oquab2023dinov2} as the sentinel teacher 
and freeze it throughout the training process. 
Knowledge derived from the remaining teachers is transferred by updating only the adapter.

\paragraph{Inference.}
After training, the encoders of all teacher models are deprecated, 
allowing the student model to operate independently in various applications. 
As the student inherits
the capabilities of all teachers, 
it can serve as a drop-in replacement for any teacher's encoder to solve tasks that the teachers are equipped to handle. 
Specifically, given the class embeddings produced by CLIP's text encoder, 
our student model enables zero-shot image classification by matching its CLIP-adapted image features. 
Similarly, the model is ready for zero-shot object detection 
by integrating GroundingDINO's decoder. 
Furthermore, \ours generates expressive image feature representations by 
complementing DINOv2's features with knowledge aggregated from the other teachers. 
Such expressive visual features allow our model to function as a conventional representation learning model, 
suitable for further training on diverse downstream tasks.

\section{Experiments}
\label{sec:exp}
\begin{table*}[t]
\centering
\adjustbox{max width=\textwidth}{
\begin{tabular}{lcccccc}
\toprule
 & \multirow{3}{*}{\shortstack[c]{Knowledge \\ Source}} &  Semantic & Instance & Object & Image\\ 
\multirow{-2}{*}{Method} & &  Segmentation & Segmentation & Detection & Classification\\ 
Metrics &    & mIoU & mAP & mAP & Top-1 Acc. & \multirow{-3}{*}{Overall}\\ 
\midrule
4M~\cite{mizrahi20244m} & data & 50.1 & 42.7 & 49.7 & 84.2 & 56.7 \\ 
GLID~\cite{liu2024glid} & data & 52.7 & - & 51.2 & - & - \\ 
Florence-2~\cite{xiao2023florence2} & data & \underline{54.9} & 42.4 & 53.6 & - & - \\ 
Uni-perceiver~\cite{zhu2022uniperceiver} & data & - & \underline{50.6} & \textbf{58.6} & \underline{86.3} & - \\ 
GroundingDINO$^{\star\dagger}$~\cite{liu2023grounding} & data & 49.7 & 48.2 & 53.9 & 79.5 & 57.8 \\ 
CLIP$^{\star\dagger}$~\cite{radford2021learning} & data & 41.5 & 33.4 & 38.0 & 75.1 & 47.0 \\ 
MAE~\cite{he2022mae} & data & 46.1 & 39.9 & 48.3 & 84.2 & 54.6 \\ 
DINOv2$^{\star\dagger}$~\cite{oquab2023dinov2} & data & 53.2 & 45.8 & 52.7 & 82.9 & 58.7 \\ 
\midrule
UNIC$^\ddagger$~\cite{sariyildiz2024unic} & model & 39.6 & - & - & 83.8 & - \\
RADIO-v2.5$^{\dagger}$~\cite{heinrich2025radiov2} & model & 51.5 & 45.1 & 51.1 & 82.7 & 57.6 \\ 
\textbf{\ours}$^{\dagger}$ (Ours) & model & 52.9 & 47.3 & 52.8 & 84.3 & \underline{59.3} \\ 
\textbf{\ours} (Ours) & model & \textbf{57.5} & \textbf{51.3} & \underline{57.7} & \textbf{86.4} & \textbf{63.2} \\ 
\bottomrule
\end{tabular}}
\caption{
Comparison to the state-of-the-art vision foundation models
on four computer vision tasks.
Existing models are 
categorized into data-centric and model-driven approaches.
The ViT-Base variant of the adopted teacher models are highlighted with $\star$.
We reproduce the results of models marked with $\dagger$
using the same task decoders as \ours with frozen image features. 
Results borrowed from the original papers are marked with $\ddagger$.
We report the mean performance for models with available results on all tasks
as an overall metric.
The 1st/2nd best results are in bold and underline, respectively.
}
\label{tab:sota}
\end{table*}
\paragraph{Datasets.}
For VFM training,
we extend the training split of ImageNet-21K~\cite{ridnik2021imagenet}
by a subset of OpenImages~\cite{benenson2022colouring},
which is initially curated for dense prediction tasks,
for enhanced diversity.
This results in $\sim 15$ million images,
which is only around $10\%$ of the training data used in DINOv2~\cite{oquab2023dinov2} and Florence-2~\cite{xiao2023florence2}.
%
In the evaluation, 
we extensively verify the effectiveness of our \ours model on four fundamental computer vision tasks requiring various levels of visual understanding from image-wise classification to region-wise detection and pixel-wise segmentation. 
For image classification, 
we evaluate on the ImageNet-1K~\cite{deng2009imagenet} dataset, 
which contains over $1.2$ million training images and $50$K validation images spanning $1,000$ diverse object categories, 
representing a broad range of visual concepts. 
For object detection and instance segmentation, we use the COCO dataset~\cite{lin2014microsoft}, 
which includes over $200$K labeled images with detailed annotations for $80$ object categories, 
emphasizing complex scenes with multiple objects. 
We assess semantic segmentation on the ADE20K dataset~\cite{zhou2017scene}
with over $20$K images annotated at the pixel level across $150$ semantic categories.
%
\paragraph{Evaluation Metrics.}
We employ standard metrics for evaluation on the four selected tasks. 
For image classification, 
we append a single linear layer to the features produced by our VFM and 
train it on ImageNet-1K using a cross-entropy loss function. 
The top-1 accuracy on the validation split is then reported. 
For object detection and instance segmentation, 
we adopt the detection decoder of Mask-DINO~\cite{li2023maskdino}
and report the mean Average Precision for bounding boxes (AP$_b$) across IoU thresholds and 
for masks (AP$_m$) as the performance metrics. 
For semantic segmentation, 
we take UPerNet~\cite{xiao2018unified} with an auxiliary Fully Convolutional Network~\cite{long2014fully} as the decoder, train it using a pixel-wise cross-entropy loss, 
and report the mean Intersection over Union (mIoU).

\paragraph{Implementation Details.}
We adopt a ViT-Base~\cite{dosovitskiy2020image} model initialized with DINOv2's pre-learned weights~\cite{dinov2repo} as the knowledge base, 
augmented by $K=4$ adapter interaction blocks, resulting in a total of approximately $100$ million parameters. 
Following our knowledge preservation strategy, 
only the adapter is updated during training,
which includes $13.6$ million trainable parameters. 
The adapter produces multi-scale feature maps to be selected and aligned with teacher's feature maps of the closest size.
The training process for \ours spans approximately $12$ hours for $40$K steps on $8\times$ H100 GPUs
while the state-of-the-art model~\cite{heinrich2025radiov2}
spends more than $14$K GPU hours for training their ViT-Base model~\cite{ranzinger2024phi}.
In each training step, 
we feed data batches of different sizes to align with different teachers through multiple forward passes. 
The losses computed from these forward passes are accumulated and used for gradient computation in a single backward pass. 
This multi-input strategy allows us to adapt batch sizes based on GPU capacity and input resolutions specified by teachers. 
In particular, we use a batch size of $32$ for knowledge transfer from CLIP~\cite{radford2021learning} 
and a batch size of $4$
for GroundingDINO~\cite{liu2023grounding}. 
%
We optimize the model using the AdamW optimizer with a learning rate of $0.0002$, 
a weight decay of $0.05$, and a cosine annealing learning rate scheduler.
For evaluation with post-training on downstream tasks,
we append the task-specific decoders altogether into our student model $f_\theta$ 
to be trained simultaneously 
for efficiency concerns.
With the multi-input strategy adopted for knowledge transfer,
the decoders are trained using their corresponding task data and labels.
The learning rate schedule and optimizer are the same as the knowledge transfer stage.
We train the model with $80$K and $200$K steps with frozen and learnable features, respectively.

\subsection{Comparison to the State-of-the-Art Models}
In \cref{tab:sota},
we conduct extensive experiments on four fundamental visual perception tasks
to compare \ours with both data-centric VFMs 
and the state-of-the-art model-driven approaches.
The competitors are selected based on their outstanding performance and versatility
with an image encoder backbone of a similar number of parameters as our \ours.

Among the three chosen teachers, 
GroundingDINO~\cite{liu2023grounding} performs exceptionally well on 
tasks which 
are specifically designed for or closely related to,
\ie, object detection and instance segmentation.
DINOv2~\cite{oquab2023dinov2} demonstrates strong performance in visual semantic understanding at both pixel and image levels, 
and its overall superior average performance across the four tasks ($58.7\%$)
highlights the advantage of general-purpose models
on generalization to various downstream tasks.
Notably, 
when training the randomly initialized decoders on top of the frozen features,
\ours achieves the best overall result,
showcasing the benefits of inheriting knowledge from multiple teachers to integrate their strengths. 
By comparing to the state-of-the-art knowledge transfer methods~\cite{heinrich2025radiov2,sariyildiz2024unic},
our model maintains non-negligible performance advantages across the board,
which demonstrates the effectiveness of our model designs for learning from diverse teachers.
By unfreezing our image features to be jointly trained with task-specific decoders\footnote{
Unfreezing teacher's features failed to yield consistent improvements,
\eg, GroundingDINO’s overall score dropped from $57.8$ to $56.9$.
We report their results using frozen features for consistency with subsequent studies.
},
\ours achieves remarkable performance gains and 
outperforms most of the strong data-centric models.
This implies the immense potential of leveraging the collective knowledge 
for building versatile VFMs.

\begin{figure}[ht]
\centering
\includegraphics[width=0.9\linewidth]{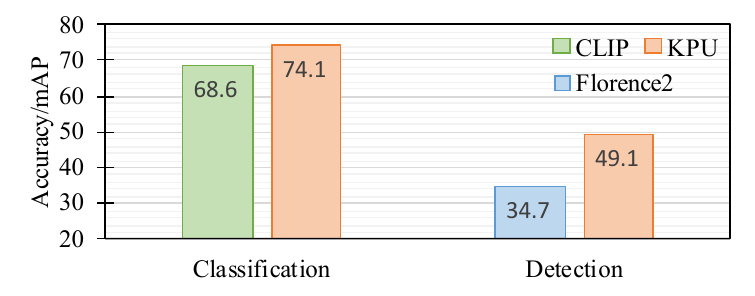}
\caption{
Comparison of zero-shot classification on ImageNet-1K and zero-shot object detection on COCO.
}
\label{fig:zsl}
\end{figure}

\paragraph{Inference with Teacher's Decoders.}
By learning to mimic teachers' behaviors within their latent spaces, 
\ours can function in various zero-shot scenarios 
as a drop-in replacement for teachers' image encoders, 
seamlessly integrating with their decoders. 
We evaluate this plug-and-play zero-shot capability by 
comparing \ours with VFMs trained with task-relevant labels,
such as CLIP~\cite{radford2021learning} trained on image-text pairs and 
Florence-2~\cite{xiao2023florence2} with bounding box labels for object detection. 
As illustrated in \cref{fig:zsl}, 
\ours achieves competitive zero-shot performance by inheriting knowledge from larger models or specialists, 
despite not being trained on any task-specific labeled data. 
These findings highlight the enhanced usability of \ours compared to conventional representation learning methods~\cite{oquab2023dinov2} 
on supporting downstream tasks without additional task-specific training. 
%

\subsection{Components Analysis}

We further conduct comprehensive component analysis to provide an in-depth understanding of our proposed designs. 
Since all the introduced components are utilized during the training of our image encoder, 
we perform a more direct evaluation of the encoder by freezing it and training only the decoder heads in all ablation experiments.

\paragraph{Effects of Knowledge Preservation.}
To investigate the benefits of the proposed knowledge preservation strategy, 
we construct a baseline model (Base$_\text{a}$)
using the ViT-Base architecture initialized with DINOv2's pre-trained weights. 
This model is trained end-to-end via feature alignment with all the three teachers~\cite{radford2021learning,liu2023grounding,oquab2023dinov2}, 
using only the $\mathcal{L}_\text{s2t}$ loss function (\cref{eq:kt_s2t}). 
Base$_\text{a}$ aligns with AM-RADIO~\cite{Ranzinger_2024_CVPR}
regarding model designs while using the same teachers as \ours.
Building upon this baseline, 
we introduce the knowledge preservation design in Base$_\text{b}$ and freeze the ViT part in our network
to train the adapter with the same objective function. 
Given the space limit,
we denote object detection, instance segmentation, semantic segmentation and image classification as Det, ISeg, SSeg and Cls, respectively, in \cref{tab:ablation} and the following.
Comparing the results of Base$_\text{a}$ and Base$_\text{b}$ in \cref{tab:ablation}, 
we observe that the model trained with our knowledge preservation strategy 
outperforms the baseline across all tasks, 
including those where the sentinel teacher excels (dense prediction tasks such as segmentation) and 
those targeted by the remaining teachers. 
These findings highlight the effectiveness of our knowledge preservation strategy in 
retaining the sentinel teacher's knowledge while 
seamlessly integrating additional knowledge from other teachers.

\begin{table}[ht]
\centering
\adjustbox{max width=\linewidth}{
\begin{tabular}{cccccccc}
\toprule
Model & Pre & Uni & Rec & Det & ISeg & SSeg & Cls\\ 
\midrule
Base$_\text{a}$ & \xmark & \xmark & \xmark & 50.3 & 44.4 & 51.1 & 81.9\\ 
Base$_\text{b}$ & \cmark & \xmark & \xmark & 52.0 & 46.8 & 52.4 & \textbf{84.4}\\ 
Base$_\text{c}$ & \cmark & \cmark & \xmark & 52.0 & 46.4 & 51.2 & 84.0\\ 
\ours & \cmark  & \cmark & \cmark & \textbf{52.9} & \textbf{47.3} & \textbf{52.8} & 84.3\\ 
\bottomrule
\end{tabular}}
\caption{
Ablation studies about 
knowledge preservation (`Pre'), 
unification (`Uni') 
and reconstruction $\mathcal{L}_\text{rec}$ (`Rec').
}
\label{tab:ablation}
\end{table}
\begin{table}[ht]
\centering
\setlength{\tabcolsep}{0.3cm}
\begin{tabular}{cccccc}
\toprule
CLIP & GDINO & Det & ISeg & SSeg & Cls\\ 
\midrule
\cmark & \xmark & 51.4 & 45.7 & \textbf{52.8} & 83.0 \\ 
\xmark & \cmark &  52.8 & 47.0 & 50.7 & 81.5 \\ 
\cmark & \cmark & \textbf{52.9} & \textbf{47.3} & \textbf{52.8} & \textbf{84.3}\\ 
\bottomrule
\end{tabular}
\caption{
Effects of knowledge transfer from multiple teachers.
GDINO is short for GroundingDINO~\cite{liu2023grounding}.
}
\label{tab:multi_teachers}
\end{table}

\paragraph{Effects of Unified Knowledge Transfer.}
%
The comparison of Base$_\text{b}$ and \ours 
reported in \cref{tab:ablation} demonstrates 
the general effectiveness of our unified knowledge transfer approach across most tasks. 
We further investigate the benefits brought by our unification design
following PHI-S~\cite{ranzinger2024phi}.
Specifically,
we measure the gap between teacher distributions
by the difference in the standard deviation of their features
on ImageNet-1K's evaluation split.
It turns out that
the standard deviation of GroundingDINO is $33.4\times$ larger than that of CLIP
in their own latent spaces,
which is reduced to $4.5\times$ after being unified in the shared space.
By shrinking the gap between teachers,
\ours mitigates the impacts of imbalanced transfer from different teachers
which is prone to a biased student.
Furthermore,
to justify the regularization $\mathcal{L}_\text{rec}$ we imposed on the common latent space,
we built an additional baseline Base$_\text{c}$ which is identical to \ours
but trained without $\{h^t_\text{rec}\}_{t=1}^T$ and the losses for reconstructing teachers' features.
Base$_\text{c}$ yields inferior performances to 
not only \ours but also
Base$_\text{b}$ trained without knowledge unification.
This indicates that our proposed regularization is indispensable
for retaining information in teachers' features
when learning the common space for knowledge unification.

\paragraph{Knowledge Transfer from Multiple Teachers.}
We further examine the importance of different teachers in knowledge transfer.
To this end, we keep all the \ours's designs
and evaluate the performance after removing either GroundingDINO~\cite{liu2023grounding} or CLIP~\cite{radford2021learning} as a teacher. 
The results in \cref{tab:multi_teachers} reaffirm the superior performance of \ours when aggregating knowledge from multiple diverse teachers
as they provide affluent knowledge from different views.
Interestingly, we observe that knowledge transfer from a single teacher does not always enhance the knowledge inherited from the sentinel teacher. 
This observation implies 
the unique challenges and advantages
of knowledge transfer from multiple teachers
lie in their diverse knowledge~\cite{tong2024eyes}.

\begin{figure}[ht]
\centering
\includegraphics[width=\linewidth]{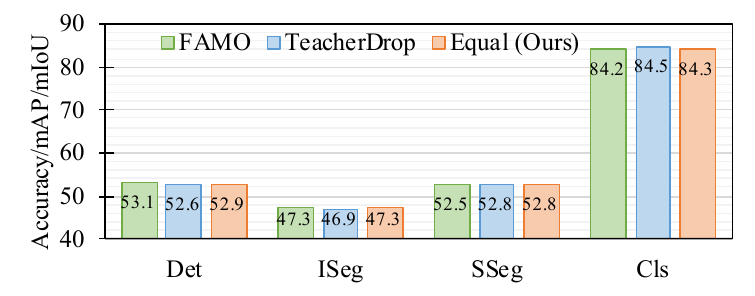}
\caption{
Effects of teachers weighting.
\ours sets the weights of teachers equally (`Equal').
`FAMO' and `TeacherDrop' adapt the weights dynamically
to balance teachers' contributions.
}
\label{fig:weighting}
\end{figure}

\paragraph{Teachers Weighting.}
In our model training,
we follow ~\citet{Ranzinger_2024_CVPR} to set the weights of feature alignment with different teachers equally.
To study the impacts of teacher weighting,
we further experiment with two different weighting strategies: 
\textit{FAMO}~\cite{bo_liu_neurips_2023} balances losses based on their gradient contributions, 
and \textit{TeacherDrop} was
specifically designed for teachers balancing~\cite{sariyildiz2024unic}.
As shown in \cref{fig:weighting}, 
the performance improvements from these advanced weighting strategies are negligible. 
While TeacherDrop is shown to be effective in multi-teacher knowledge transfer~\cite{sariyildiz2024unic},
the marginal improvement it brings to \ours 
implies the potentially less severe task/teacher interference 
thanks to our knowledge unification design.

\section{Conclusion}
In this work, we build a VFM that can generalize well across diverse downstream tasks by 
inheriting condensed visual knowledge from powerful pre-trained models. 
This is accomplished by our proposed
\Ours (\ours), a simple yet effective model-driven training method for VFMs. 
\ours unifies teachers in a common latent space
for better exploration of their complementary insights.
At the same time,
it explicitly preserves the general knowledge from a sentinel teacher to benefit from its remarkable generalization ability
and facilitate effective training on downstream tasks.
By deriving the collective knowledge of diverse teachers with unique visual perspectives, 
\ours not only produces generalizable visual features
but also supports zero-shot image classification and object detection
without the need for further tuning with category or bounding-box labels.
\ours yields superior performance on multiple fundamental vision tasks over existing data-centric and knowledge transfer approaches,
unlocking 
the potential of model-driven training for building strong VFMs.

\bibliography{aaai2026}

\clearpage
\appendix

\end{document}